\documentclass{article}

\usepackage{arxiv}

\usepackage{todonotes}
\usepackage{booktabs}
\usepackage{balance}
\usepackage{amssymb}
\usepackage{pifont}
\usepackage[section]{placeins}
\usepackage{makecell}
\usepackage{float}
\usepackage{todonotes}
\usepackage{amsmath}
\usepackage{graphicx}
\graphicspath{{./images/}} 

\begin{document}
%
\title{Accelerating Convolutional Neural Network Pruning via Spatial Aura Entropy}

\author{
 Mu\c sat Bogdan\\
  Transilvania University of Bra\c sov, Romania, \\ 
  Department of Electrical Engineering and Computer Science, \\
  \texttt{bogdan\_musat\_adrian@yahoo.com} \\
   \And
 Andonie R\u azvan \\
  Central Washington University, USA, \\ Department of Computer Science \\
  \texttt{razvan.andonie@cwu.edu} \\
}

\maketitle

\begin{abstract}
In recent years, pruning has emerged as a popular technique to reduce the computational complexity and memory footprint of Convolutional Neural Network (CNN) models. Mutual Information (MI) has been widely used as a criterion for identifying unimportant filters to prune. However, existing methods for MI computation suffer from high computational cost and sensitivity to noise, leading to suboptimal pruning performance. We propose a novel method to improve MI computation for CNN pruning, using the spatial aura entropy. The spatial aura entropy is useful for evaluating the heterogeneity in the distribution of the neural activations over a neighborhood, providing information about local features. Our method effectively improves the MI computation for CNN pruning, leading to more robust and efficient pruning. Experimental results on the CIFAR-10 benchmark dataset demonstrate the superiority of our approach in terms of pruning performance and computational efficiency.

\end{abstract}



%

\section{Introduction}

Deep Convolutional Neural Networks (CNNs) have achieved state-of-the-art performance on a wide range of computer vision tasks, such as image classification, object detection, and semantic segmentation \cite{krizhevsky2012imagenet,he2016deep,long2015fully}. These models typically consist of a large number of parameters, making them computationally expensive and memory-intensive to deploy on resource-limited devices, such as mobile phones and embedded systems. Training these models requires significant computational resources and time, which limits the ability to explore large-scale architectures and hyperparameters \cite{sun2019patient}.

One promising approach to alleviate these challenges is pruning, which refers to the process of reducing the size of a neural network by removing unimportant weights, neurons, or filters, without significant loss in accuracy \cite{lecun1990optimal}. Pruning can result in more efficient models that require fewer parameters, consume less memory, and have faster inference time. This can be particularly important for real-time applications, where latency and energy consumption are critical factors \cite{wu2020survey}.

Pruning techniques can be classified into two main categories: weight pruning and structured pruning. Weight pruning involves setting small weights to zero, which can lead to sparse connectivity patterns \cite{han2015learning}. On the other hand, structured pruning removes entire neurons or filters, which preserves the dense connectivity patterns but reduces the model size \cite{li2017pruning}. Several recent works have explored different pruning methods, such as magnitude-based pruning \cite{han2015learning}, sensitivity-based pruning \cite{molchanov2019importance}, and filter pruning \cite{li2017pruning}, among others. These techniques can be applied during or after training and can be combined with other compression methods, such as quantization, to further reduce the model size \cite{han2016eie}.

Despite the potential benefits of pruning, there are also some challenges that need to be addressed. For example, pruning can lead to a significant increase in the number of training iterations required to recover the accuracy of the original model, which can offset the benefits of the reduced model size \cite{liu2018rethinking}. Moreover, the choice of the pruning method, the pruning rate, and the fine-tuning strategy can affect the final accuracy and the efficiency of the pruned model \cite{frankle2019lottery}. Therefore, it is essential to carefully design and evaluate pruning methods for different applications and network architectures.

Sarvani \emph{et al.} \cite{SARVANI2022186} introduced an Information Bottleneck theory \cite{tishby_information_1999} based filter pruning method that utilizes Mutual Information (MI) to determine filter significance. Their method outperformed recent state-of-the-art pruning methods and has been demonstrated on LeNet-5, VGG-16, ResNet-56, ResNet-110, and ResNet-50 architectures using MNIST, CIFAR-10, and ImageNet datasets. However, the  kernel-based MI estimation used in \cite{SARVANI2022186} has several computational issues, including kernel width selection and curse of dimensionality. These issues can result in an inaccurate estimation of MI and limit the practical application of kernel-based methods. This gave us the motivation for our current research.

In this paper, we build on top of Sarvani \emph{et al.}'s work and propose a more efficient method to compute the MI required for filter importance selection. In our experiments, we reduced the execution time from almost a week of to a single day. Our method computes MI using the spatial aura entropy, as defined in our previous work \cite{Musat2022}. The spatial aura entropy method is more efficient and straightforward than kernel-based estimators and does not require the selection of kernel width. Our method manages to preserve or improve the results obtained by the original work, but at a much lower computational cost. 

Visualization can be a useful tool for comparing various pruning techniques by showcasing how they affect the model's prediction accuracy. By providing a visual representation of these impacts, users can gain a better understanding of the effectiveness of different pruning strategies. This information can be used to refine pruning strategies and improve the overall effectiveness of the pruning process \cite{Li2022}. When it comes to pruning, visualization can integrate the benefits of artificial intelligence, machine learning, and visual analytics \cite{Kovalerchuk2022, Kovalerchuk2022integrating}. Our experiments involve the visualization of the pruning process, allowing us to gain a better understanding of the impact of the pruning on the model's performance.

The rest of the paper is structured as follows. Section \ref{related} presents related work. Section \ref{background} summarizes how we compute MI using the spatial aura entropy. In Section \ref{method} we introduce our novel pruning method and in Section \ref{experiments} we describe experimental results. Section \ref{conclusion} contains the final remarks.

\section{Related Work} \label{related}

CNNs have become the de facto standard for many computer vision tasks, thanks to their impressive performance. However, CNNs are often computationally expensive and memory-intensive, which limits their deployment on resource-constrained devices such as mobile phones and embedded systems. Therefore, various pruning techniques have been proposed to reduce the complexity of CNNs, while maintaining their accuracy. In this section, we discuss some of the most popular CNN pruning techniques.

The lottery ticket hypothesis proposed by Frankle and Carbin~\cite{frankle2019lottery} suggests that there exists a sparse subnetwork within an over-parameterized network, which can be trained to achieve the same accuracy as the original network. The lottery ticket hypothesis is based on the idea of iterative pruning and retraining: a network is pruned, and the remaining weights are fine-tuned to recover the accuracy. This process is repeated until the desired level of sparsity is achieved. The authors showed that by using their method, the number of parameters in a network can be reduced by up to 90\% without loss of accuracy.

Another popular pruning technique is magnitude pruning~\cite{han2015learning}, which involves pruning the weights with the smallest magnitude. This method is simple and can be applied to any neural network, including CNNs. Han \emph{et al.} showed that by pruning 90\% of the weights, the computational cost of a CNN can be reduced by a factor of 10 with only a slight drop in accuracy.

Structured pruning is another pruning technique that aims to remove entire filters, channels or even layers from the network. A popular structured pruning method is channel pruning~\cite{he2017channel}, where entire channels are pruned based on their importance, which is typically measured by their L1-norm. Channel pruning is particularly effective in reducing the computation cost of a network, as it removes entire channels and their associated computations. Another structured pruning method is filter pruning~\cite{li2016pruning}, where entire filters are pruned based on their importance. Filter pruning is more aggressive than channel pruning, as it removes entire filters and their associated weights, which results in a more compact network. Ultimately, as demonstrated in one of our prior studies \cite{e22121365}, it is feasible to perform even whole layer pruning. Specifically, layers can be removed if the spatial entropy of the saliency maps remains unchanged from the previous layer.

The basis for our work is the paper of Sarvani \emph{et al.} \cite{SARVANI2022186}, which uses a novel criterion for structured pruning. The method involves computing MI to identify the most crucial feature maps in each layer and removing filters associated with unimportant feature maps. High Relevance Filters (HRel) filters (filters exhibiting higher MI when considering class labels) are retained. This method (we will refer to it as the \textbf{HRel method}) uses a matrix-based R\'enyi's alpha entropy estimator \cite{wickstrom2019information} to approximate MI. This estimation technique incurs a significant computational cost due to the need for continuously updating the bandwidth required for the kernels used in MI estimation.    

Several other pruning methods were proposed, including weight-sharing pruning~\cite{liu2018rethinking}, which shares weights among different neurons to reduce the number of unique weights, and morphological pruning~\cite{gordon2018morphnet}, which learns the topology of the network in a resource-constrained manner. The Once-for-All method  \cite{han2021once}, trains a large neural network that contains many subnetworks, sharing weights and searches among them to find the most efficient one for a specific target device.

Recently, there have been efforts to combine pruning with other techniques, such as quantization and knowledge distillation. For instance, Liu \emph{et al.} \cite{liu2020autocompress} introduced an automated framework for neural network compression, which combines pruning with other compression techniques such as quantization and low-rank factorization. Similarly, Cao \emph{et al.} \cite{cao2019learning} proposed a sparsity regularization technique that encourages the network to learn a compression-agnostic representation.

In addition to these methods, there have been efforts to design CNN architectures that are inherently more efficient, requiring fewer parameters and less computation for inference. One such architecture is MobileNet \cite{howard2017mobilenets}, which is designed for mobile devices with limited computing resources. Another is EfficientNet \cite{tan2019efficientnet}, which uses a compound scaling method to optimize network depth, width, and resolution to achieve state-of-the-art accuracy with significantly fewer parameters.

\section{Background: Mutual Information Computation using Spatial Entropy} \label{background}

To make the paper self-contained, we summarize in this section the MI estimation technique used for quantifying the importance of  convolutional filters. The MI estimation technique used in this work is based on the formulation of MI as defined in  \cite{Musat2022}, and uses spatial aura entropy estimation. 

As in \cite{SARVANI2022186}, we compute the MI between each activation map and the one-hot encoded ground truths. The MI between two random variables $X$ and $Y$ is computed as:

\begin{equation} \label{formula:mi}
    I(X, Y) = H(X) + H(Y) - H(X, Y)
\end{equation}

\noindent
where $H(\cdot)$ represents the entropy of a variable, and $H(\cdot, \cdot)$ represents the joint entropy between two variables. Since convolutional feature maps have embedded into them the notion of spatiality, the univariate formula of the Shannon entropy does not take into consideration this special property. As such we make use of the spatial aura matrix entropy, as initially defined in \cite{Volden1995}.  

We define the joint probability of two features cells at spatial locations $(i, j)$ and $(i + k,\:j + l)$ to take the values $g$, respectively $g'$ as:

\begin{equation} \label{formula:2}
    p_{gg'}(k, l) = P(X_{i,\:j} = g, X_{i + k,\:j + l} = g')
\end{equation}
where $g$ and $g'$ are discretized variables, obtained after binning the values of the action maps. The probabilities are computed by dividing the number of occurrences for $(g, g')$ by the total number of possible occurrences. To reduce the computational overhead, we assume that $p_{gg'}$ is independent of $(i, \:j)$ (the homogeneity assumption \cite{Journel1993}), and we define for each pair $(k,\:l)$ the entropy:

\begin{equation}
    H(k,\:l) = - \sum_{g} \sum_{g'} p_{gg'} (k,\:l) \log p_{gg'} (k,\:l)
\end{equation}
where the summations are over the number of possible binned values (128 in our case). A standardized relative measure of bivariate entropy is \cite{Journel1993}:

\begin{equation} \label{eq:4}
    H_R (k,\:l) = \frac{H(k,\:l) - H(0)}{H(0)} \in [0,\:1]
\end{equation}

The maximum entropy $H_R (k,\:l)=1$ corresponds to the case of two independent variables. $H(0)$ is the univariate entropy, which assumes all feature cells as being independent, and we have $H(k,\:l) \geq H(0)$. 

Based on the relative entropy for $(k,\:l)$, the Spatial Disorder Entropy (SDE) for an $m \times n$ image $\textbf{X}$ was defined in \cite{Journel1993} as:

\begin{equation} \label{eq:5}
    H_{SDE} (\textbf X) \approx \frac{1}{mn} \sum_{i = 1}^m \sum_{j = 1}^n \sum_{k = 1}^m \sum_{l = 1}^n H_R (i - k,\:j - l)
\end{equation}

Since the complexity of SDE computation is high, we decided to use a simplified version - the Aura Matrix Entropy (AME, see \cite{Volden1995}), which only considers the second order neighbors from the SDE computation:

\begin{align} \label{eq:6}
\begin{split}
    H_{AME} (\textbf X) \approx \frac{1}{4} \bigg( H_R (-1,\:0) + H_R (0,\:-1) \\ 
    + H_R (1,\:0) + H_R (0,\:1) \bigg)
\end{split}
\end{align}

Putting it all together, starting from a discretized feature map, we compute the probabilities $p_{gg'}$ in equation (\ref{formula:2}), and finally the AME in equation (\ref{eq:6}), which results in the spatial entropy quantity of the activation map $X$.

The entropy of Y is guaranteed to be $0$ because Y is encoded using one-hot encoding, which places all the magnitude on a single position. As such, in our case, the MI formula becomes: 

\begin{equation} \label{formula:mi_new}
    I(X, Y) = H(X) - H(X, Y)
\end{equation}

We modify the simplified aura matrix entropy to be applicable for joint entropy calculation by changing equation (\ref{formula:2}) to:

\begin{equation}
    \begin{aligned} \label{formula:mi_pg}
        p_{gg'g^{''}g^{'''}}(k, l) = P(X_{i,\:j} = g, X_{i + k,\:j + l} = g', \\
        Y_{i,\:j} = g^{''}, Y_{i + k,\:j + l} = g^{'''})
    \end{aligned}
\end{equation}

\noindent
where $g$, $g'$, $g^{''}$, $g^{'''}$ are again discretized values. The upcoming equations from the spatial entropy computation will use $p_{gg'g^{''}g^{'''}}$ instead of $p_{gg'}$. The final modification will be to equation (\ref{eq:6}), where we take into consideration four spatial positions instead of two, the first two from $X$ and the last two from $Y$: 

\begin{equation} \label{eq:9}
    \begin{aligned}
        H_{AME} (\textbf X, \textbf Y) \approx \frac{1}{4} \bigg( & H_R (-1,\:0, -1,\:0) + H_R (0,\:-1, 0,\:-1)
        \\ & + H_R (1,\:0, 1,\:0) + H_R (0,\:1, 0,\:1) \bigg)
    \end{aligned}
\end{equation}

As $Y$ is a one-dimensional vector with $c$ (classes) one-hot encoded values, and an activation map $X$ is typically a $3D$ tensor $(C\times W\times H)$, it is necessary to establish a common spatial dimensionality between the two variables in order to compute the values in equation (\ref{formula:mi_pg}). To achieve this, we first repeat the values of $Y$ for $c$ times to create a $c\times c$ grid and convert it into a $2D$ matrix. We then use interpolation with constant values to scale the newly obtained matrix to the same spatial dimensionality as $X (W\times H)$. Once both variables share the same spatial size, we can calculate the values in equation (\ref{formula:mi_pg}) by inputting each channel of $X$ with the transformed $Y$ matrix. Finally, we use equation (\ref{eq:9}) to determine the joint spatial entropy between $X$ and $Y$.

In order to calculate the value of equation (\ref{formula:mi_new}), we utilize the outcome of equation (\ref{eq:6}) for the entropy of $X$, as well as equation (\ref{eq:5}) for the joint entropy of $X$ and $Y$.

\section{Our Method} \label{method}

We introduce in this section our CNN filter pruning method. Primarily, it follows the workflow of the HRel method. 

First, we train a CNN until convergence. Then we compute the MI between the activation maps and the ground truth labels for each layer in the network. The activation maps represent the output of each filter in the layer when the input is passed through the layer.

Next, we rank the filters in descending order of MI and remove a certain percentage of filters with the lowest MI. After the filters are removed, we perform a fine-tuning step to recover the network's original performance. Fine-tuning involves training the pruned network for some epochs using a smaller learning rate than the original training rate.

Once the fine-tuning step is complete, we repeat the process of computing MI, ranking filters, and MI. This process continues until the maximum number of filters removed for each layer is achieved.

The advantage of using MI as a criterion for pruning is that it considers the information content of each filter in the network, rather than just its weight magnitude or gradient. Additionally, MI can capture the interactions between filters and their contribution to the overall performance of the network.

In the HRel method, MI is estimated using R\'enyi’s alpha entropy estimator \cite{wickstrom2019information}, which relies on kernel functions and is highly sensitive to the optimal kernel bandwidth of the dataset \cite{DBLP:journals/corr/abs-2005-07783}, as noted by the authors themselves. This sensitivity results in a computationally expensive procedure, as the kernel bandwidth needs to be updated continuously during training and fine-tuning, making the pruning mechanism burdensome. As such, pruning a network using the public code that the authors provided takes almost a week for a VGG-16 network \cite{simonyan2015very}, rendering it impractical for real-life applications.

In contrast to the original HRel technique, our method employs a different approach by using a MI estimation that does not rely on kernel functions. 
 Instead, we estimate MI using the spatial aura entropy (the AME simplified version)  described in Section \ref{background}. This allows for efficient computation of MI using as few as 100 samples. We eliminate the computational burden of continuously updating the kernel bandwidth during training and fine-tuning, making it significantly faster and more practical for real-life applications. Our method still preserves or even improves upon the results achieved by the original HRel paper, demonstrating its effectiveness in efficient filter selection.

\section{Experiments} \label{experiments}

This section presents empirical proof of the effectiveness of our approach. We evaluate our method on the widely used CIFAR-10 benchmark dataset \cite{cifar10}. We modified the publicly available code provided by the HRel authors \cite{SARVANI2022186} altering the MI estimation procedure. Although we observed some variations in the baseline performance of ResNet architectures from what was reported in the original paper, we compare our method and HRel employing on the same starting baseline performance.

\subsection{VGG-16}

To evaluate the effectiveness of our proposed method, we apply it to the popular VGG-16 architecture \cite{simonyan2015very}, which consists of 13 convolutional layers and two fully connected layers. We prune filters from the convolutional layers and train the network for 300 epochs with an initial learning rate of 0.1, which we reduce by a factor of 10 at epoch numbers 80, 140, and 230, until the baseline accuracy is achieved. Subsequently, we prune and retrain the network for 90 epochs with a learning rate of 0.01, which we reduce by a factor of 10 at epochs 40 and 70.

Table \ref{tab:vgg16_res} presents a comparison of the accuracy achieved by the original HRel formulation and our proposed method for different configurations of the remaining number of filters per layer on the CIFAR-10 test set. Our method outperforms HRel for both configurations, demonstrating its effectiveness in the pruning process. Specifically, the original VGG-16 network achieves a test accuracy of 93.95\%, while our proposed method yields test accuracies of 93.22\% and 93.4\% for Configuration 1 and Configuration 2, respectively. These results highlight the potential of our method in enhancing model efficiency and accuracy.

\begin{table}[htbp]
\caption{Comparison of accuracy achieved by HRel and our method on the CIFAR-10 test set across different pruning configurations. The table shows that our method outperforms HRel for both configurations, demonstrating its effectiveness in the pruning process.}
\centering
\begin{tabular}{|c|c|c|}
\hline
 & \textbf{HRel} & \textbf{Our Method} \\
\hline
Original network: &  &  \\
64-64-128-128-256-256-256 & 93.95\% & 93.95\% \\
-512-512-512-512-512-512 &  &  \\
\hline
Configuration 1: &  &  \\
19-48-64-64-95-107-107 & 93.15\% & 93.22\% \\
-175-71-71-44-44-56 &  &  \\
\hline
Configuration 2: &  &  \\
24-40-64-77-176-134-120 & 93.22\% & 93.4\% \\
-141-56-56-56-56-56 &  &  \\
\hline
\end{tabular}
\label{tab:vgg16_res}
\end{table}

\subsection{ResNet-56}

ResNet-56 \cite{he2016deep} is a more complex and deeper neural network architecture compared to VGG-16. It consists of 55 convolutional layers and 1 fully connected layer, with all convolutional layers (except the first one) grouped into three blocks, each containing 18 convolutional layers. The first, second, and third blocks have 16, 32, and 64 filters, respectively. To achieve the baseline accuracy, the network is trained for 180 epochs with an initial learning rate of 0.1, which is then decreased by a factor of 10 at epoch numbers 91 and 136.

After pruning, the network is retrained for 200 epochs with a learning rate of 0.01, which is then decreased by a factor of 10 at epochs 100 and 150. We observe that the ResNet-56 architecture requires twice the number of fine-tuning epochs using our method compared to the original HRel framework. However, the additional computational time is negligible when compared to the overall runtime of the pruning process from the original HRel.

After pruning, the final remaining number of filters in the convolutional layers of each block are 8, 15, and 30, respectively. Table \ref{tab:resnet56_res} shows that our method achieves the same accuracy as HRel. Specifically, the original HRel framework achieves an accuracy of 92.74\% with a filter configuration of 8-15-30, while our method achieves an accuracy of 92.76\% with the same configuration.

\begin{table}[htbp]
\caption{Comparison of accuracy achieved by HRel and our method on the CIFAR-10 test set using the ResNet-56 architecture with the filter configuration of 8-15-30 after pruning.}
\centering
\begin{tabular}{|c|c|c|}
\hline
 & \textbf{HRel} & \textbf{Our Method} \\
\hline
Original network: & 93.45\% & 93.45\% \\
 16-32-64 & &  \\
\hline
Configuration: & 92.74\% & 92.76\% \\
8-15-30 &  &  \\
\hline
\end{tabular}
\label{tab:resnet56_res}
\end{table}

\subsection{ResNet-110}

ResNet-110  \cite{he2016deep} is a deep neural network architecture composed of 109 convolutional layers and a single fully connected layer. The structure of ResNet-110 is similar to that of ResNet-56, where the convolutional layers are grouped into three blocks, except for the first convolutional layer. However, in ResNet-110, each block contains 36 convolutional layers with 16, 32, and 64 filters, respectively.

To achieve the baseline accuracy, the network is trained for 240 epochs with an initial learning rate of 0.1, which is decreased by a factor of 10 at epoch numbers 88, 160, and 190. The first convolutional layer is not pruned, similar to other pruning methods. After pruning, the network is fine-tuned for 70 epochs with a learning rate of 0.01, which is decreased by a factor of 10 at epochs 30 and 50. The number of remaining filters in each block's convolutional layer is 8, 15, and 30, respectively, after pruning.

Table \ref{tab:resnet110_res} presents a comparison of the accuracy achieved by our proposed method and HRel on the CIFAR-10 test set using the ResNet-110 architecture with a filter configuration of 8-15-30 after pruning. The original network with a filter configuration of 16-32-64 achieved an accuracy of 93.27\%. Our method achieved a high accuracy of 92.42\%, which is slightly improved compared to HRel's accuracy of 92.36\%, with a difference of 0.06\%. 

\begin{table}[h]
\caption{Comparison between the accuracy achieved by HRel and our proposed method on the CIFAR-10 test set using ResNet-110 architecture with a filter configuration of 8-15-30 after pruning.}
\centering
\begin{tabular}{|c|c|c|}
\hline
 & \textbf{HRel} & \textbf{Our Method} \\
\hline
Original network: & 93.27\% & 93.27\% \\
 16-32-64 & &  \\
\hline
Configuration: & 92.36\% & 92.42\% \\
8-15-30 &  &  \\
\hline
\end{tabular}
\label{tab:resnet110_res}
\end{table}

\subsection{Visualization of the pruning process}
Tables \ref{tab:conv6} and \ref{tab:conv12} display a graphical representation of the MI distributions for both methods, allowing for a visual comparison. The two distributions exhibit a comparable shape, albeit with distinct value ranges resulting from the distinct MI computation methods employed. The visualization underscores that while our approach differs from the original HRel, it still captures the pertinent information necessary for filter pruning selection.

\begin{table}[h]
\caption{MI distributions of VGG-16 for convolutional layer 5 during pruning: 256, 150, 101 filters. Top row is for original HRel, bottom row is for ours}
\centering
\setlength{\tabcolsep}{0pt}
\renewcommand{\arraystretch}{0}
\begin{tabular}{ccc}
\includegraphics[width=0.35\linewidth]{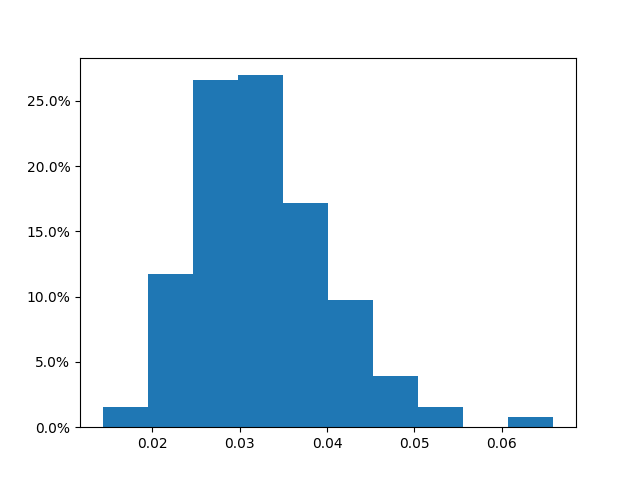} & 
\includegraphics[width=0.35\linewidth]{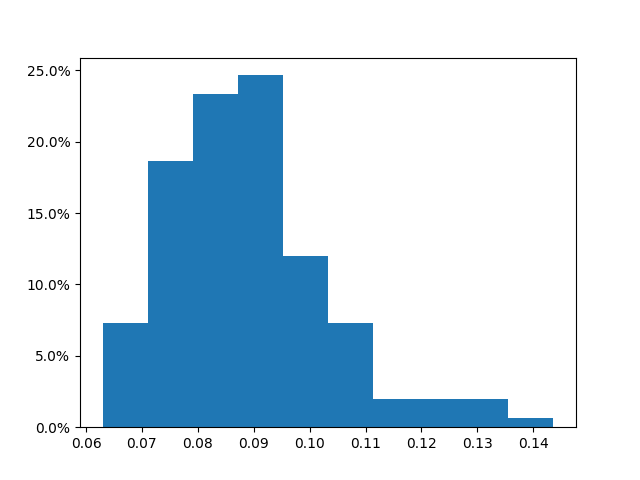} &
\includegraphics[width=0.35\linewidth]{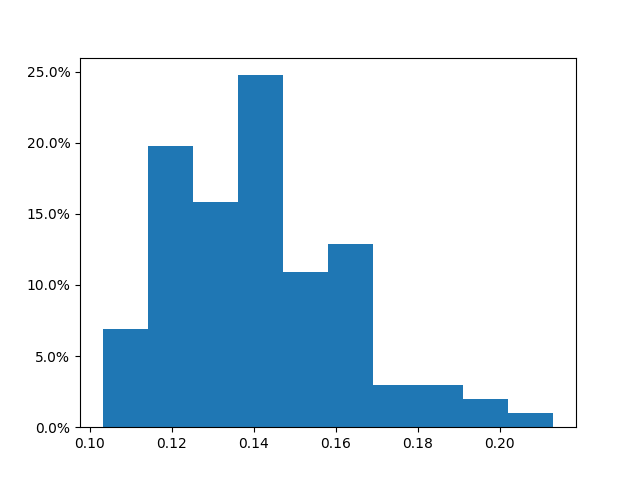} \\
\includegraphics[width=0.35\linewidth]{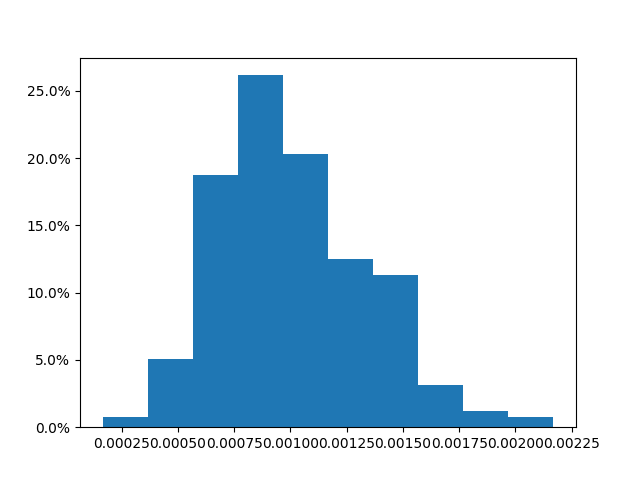} &
\includegraphics[width=0.35\linewidth]{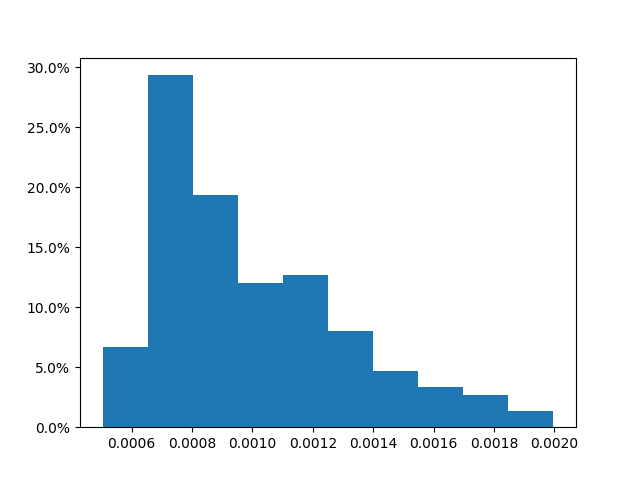} &
\includegraphics[width=0.35\linewidth]{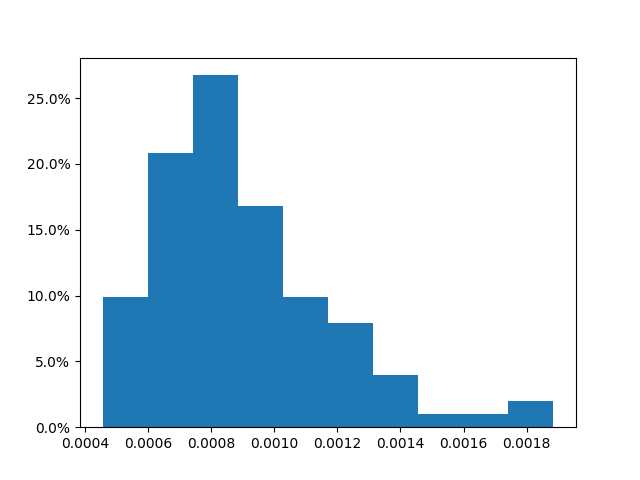} \\
\end{tabular}
\label{tab:conv6}
\end{table}

\begin{table}[h]
\caption{MI distributions of VGG-16 for convolutional layer 12 during pruning: 512, 175, 71 filters. Top row is for original HRel, bottom row is for ours}
\centering
\setlength{\tabcolsep}{0pt}
\renewcommand{\arraystretch}{0}
\begin{tabular}{ccc}
\includegraphics[width=0.35\linewidth]{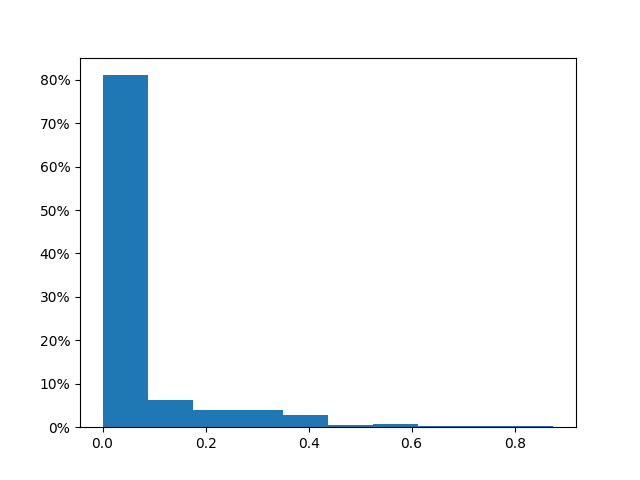} & 
\includegraphics[width=0.35\linewidth]{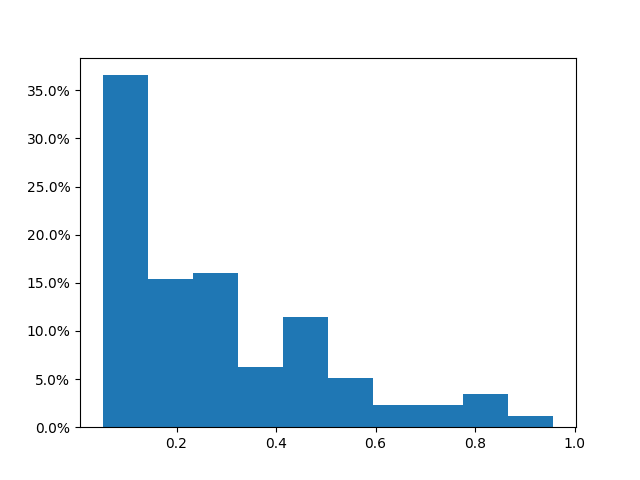} &
\includegraphics[width=0.35\linewidth]{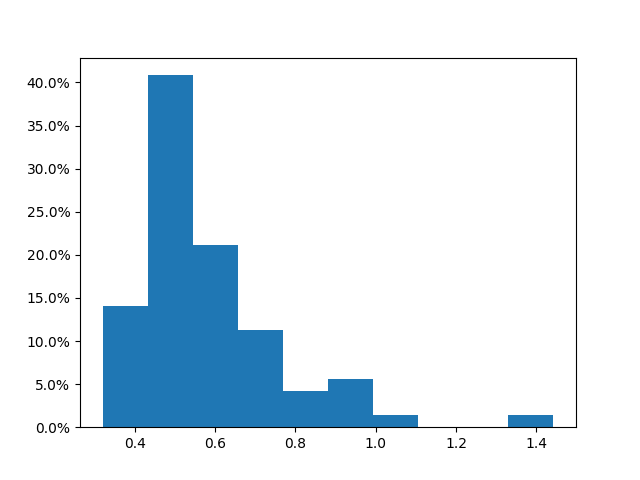} \\ 
\includegraphics[width=0.35\linewidth]{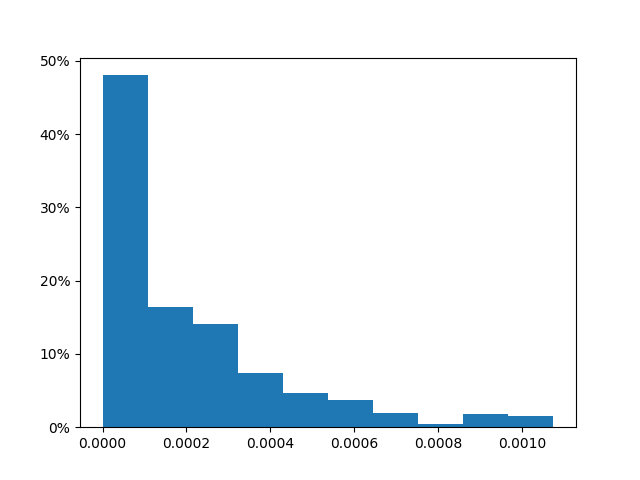} &
 \includegraphics[width=0.35\linewidth]{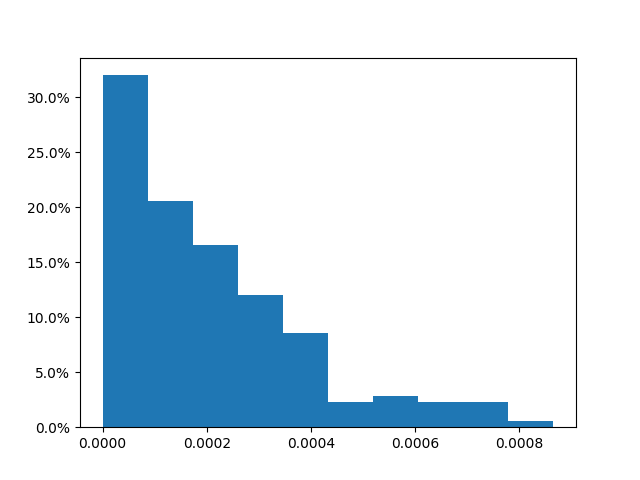} &
\includegraphics[width=0.35\linewidth]{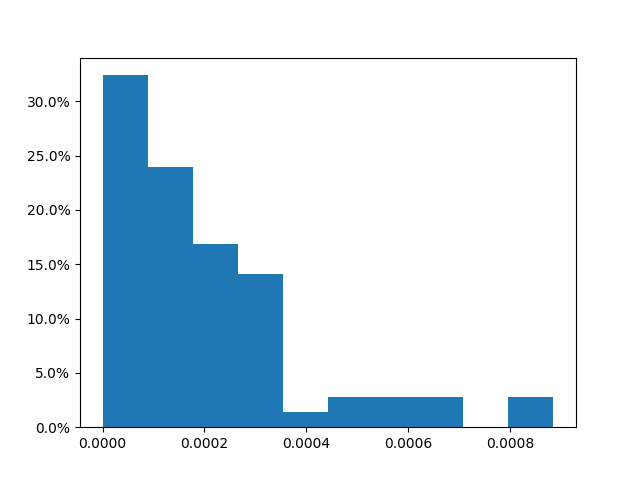} \\
\end{tabular}
\label{tab:conv12}
\end{table}
\subsection{Discussion}
The experiments demonstrate the effectiveness of our method in improving the accuracy and efficiency of CNN pruning. The incorporation of spatial aura entropy into MI calculation provides a more robust and efficient pruning method. This is achieved by improving the accuracy of filter importance selection criteria and reducing the optimization time required for MI computation. Our method not only outperforms the baseline HRel method in terms of pruning performance but also significantly reduces the computational cost, making it a practical and scalable solution for deep learning model compression.

\section{Conclusion and Future Work}\label{conclusion}
We introduced an alternative solution to the matrix-based R\'enyi's alpha entropy estimator used in the HRel method proposed in \cite{SARVANI2022186}. This improvement significantly reduces the optimization time from almost a week to a single day, making it a more practical and efficient method for large-scale model pruning. Our method is an efficient and effective solution to reducing the computational complexity and memory footprint of deep learning models, providing a viable alternative to existing methods with improved pruning performance and computational efficiency. 

Future work could focus on exploring the applicability of our method on other benchmark datasets and model architectures, as well as investigating its potential in other areas of machine learning and computer vision.

\balance
\bibliographystyle{IEEEtran}
\bibliography{bibliography}

\end{document}